\documentclass[a4paper,fleqn]{cas-dc}

\usepackage{svg}
\usepackage[numbers]{natbib}
\usepackage{float}
\usepackage{amsmath}
\usepackage{multirow}
\usepackage{pgfplots}
\usepackage{tikz}
\usepackage{graphicx}
\usepackage{booktabs}
\usepackage{amsfonts}
\usepackage{nicefrac}
\usepackage{microtype}
\usepackage{mathrsfs}
\usepackage{setspace}
\usepackage{color, colortbl}
\usepackage{marginnote}
\usepackage{todonotes}
\usepackage{fixltx2e}
\usepackage{placeins}

\pgfplotsset{compat=1.18}

\def\tsc#1{\csdef{#1}{\textsc{\lowercase{#1}}\xspace}}
\tsc{WGM}
\tsc{QE}

\ExplSyntaxOn
\cs_gset:Npn \__first_footerline:
  { \group_begin: \small \sffamily \__short_authors: \group_end: }
\ExplSyntaxOff

\begin{document}
\let\WriteBookmarks\relax
\def\floatpagepagefraction{1}
\def\textpagefraction{.001}

\title [mode = title]{SPARSE Data, Rich Results: Few-Shot Semi-Supervised Learning via Class-Conditioned Image Translation}  
\shorttitle{}


\author[1,2]{Guido Manni}
\ead{guido.manni@unicampus.it}
\author[2]{Clemente Lauretti}
\ead{c.lauretti@unicampus.it}
\author[2]{Loredana Zollo}
\ead{l.zollo@unicampus.it}
\author[1]{Paolo Soda}
\ead{p.soda@unicampus.it}

\affiliation[1]{organization={Unit of Artificial Intelligence and Computer Systems, Department of Engineering, Università Campus Bio-Medico di Roma, Rome, Italy}}

\affiliation[2]{organization={Unit of Advanced Robotics and Human-Centred Technologies, Department of Engineering, Università Campus Bio-Medico di Roma, Rome, Italy}}

\shortauthors{Guido Manni et~al.}

\begin{abstract}
Deep learning has revolutionized medical imaging, but its effectiveness is severely limited by insufficient labeled training data. This paper introduces a novel GAN-based semi-supervised learning framework specifically designed for low labeled-data regimes, evaluated across settings with 5 to 50 labeled samples per class. Our approach integrates three specialized neural networks—a generator for class-conditioned image translation, a discriminator for authenticity assessment and classification, and a dedicated classifier—within a three-phase training framework. The method alternates between supervised training on limited labeled data and unsupervised learning that leverages abundant unlabeled images through image-to-image translation rather than generation from noise. We employ ensemble-based pseudo-labeling that combines confidence-weighted predictions from the discriminator and classifier with temporal consistency through exponential moving averaging, enabling reliable label estimation for unlabeled data. Comprehensive evaluation across eleven MedMNIST datasets demonstrates that our approach achieves statistically significant improvements over six state-of-the-art GAN-based semi-supervised methods, with particularly strong performance in the extreme 5-shot setting where the scarcity of labeled data is most challenging. The framework maintains its superiority across all evaluated settings (5, 10, 20, and 50 shots per class). Our approach offers a practical solution for medical imaging applications where annotation costs are prohibitive, enabling robust classification performance even with minimal labeled data. Code is available at https://github.com/GuidoManni/SPARSE.
\end{abstract}




\maketitle

\section{Introduction}
Deep learning has demonstrated remarkable potential in revolutionizing medical imaging \cite{CHEN2022102444}.
However, insufficient labeled data for model training is one of the main challenge hindering its effectiveness that arises from several constraints such as: the stringent privacy regulations and ethical guidelines governing patient data access and distribution \cite{privacy}, and the necessity of specialized medical expertise for data annotation, a resource limited by healthcare professionals' primary commitment to patient care \cite{efficiency}.
These constraints often results in what is known as the low-data regime - a situation where the number of labeled medical images falls below the threshold needed for reliable convergence of deep networks, typically ranging from dozens to a few hundred samples depending on the complexity of the task and model architecture.
To address this issue, researchers have explored unsupervised, supervised and semi-supervised learning. \\
The approaches in the first category tackles the low-data regime through unsupervised learning, which exploits the abundant unlabeled medical images to learn meaningful representations. 
Models are trained to capture underlying data patterns through tasks like image reconstruction, anomaly detection, or feature learning. 
This approach is particularly valuable in medical imaging where unlabeled data is available, but a fundamental challenge remains: without sufficient labeled validation data, it is difficult to ensure that the extracted features are clinically relevant rather than merely statistically significant in the data distribution.
In the second case of supervised learning, researchers have explored various strategies to mitigate the limited availability of labeled samples, such as transfer learning, data augmentation and synthetic data generation. 
Transfer learning leverages models pre-trained on large datasets by fine-tuning them for specific tasks with limited data~\cite{TransferLearning}, but its effectiveness diminishes when the target domain differs significantly from the source domain \cite{TransferLearningCon}. 
Data augmentation techniques artificially expand training datasets through various transformations~\cite{DA}, but cannot introduce new information.
Synthetic data generation through simulation or generative models has emerged as another strategy~\cite{Synthetic_1, Synthetic_2}, though generating fully synthetic datasets that realistically capture real-world data distributions remains challenging~\cite{Synthetic_con_1, Synthetic_con_2}.
The third approach is  semi supervised learning (SSL) that simultaneously leverages labeled and unlabeled data.
Traditional SSL methods relied on  machine learning techniques such as self-training and co-training.
However, recent advances in generative models, particularly Generative Adversarial Networks (GANs) based methods, have improved SSL performance, not by generating completely synthetic data but by learning to extract meaningful features from the real unlabeled data distribution to enhance the learning process \cite{SSL_Surv}.
Despite these advances, existing SSL methods often struggle in low-data regime as it happens in  medical imaging. 
In this paper we tackle this issue and we introduce the following contributions:
\begin{itemize}
    \item A novel GAN-based semi-supervised learning method specifically designed for medical image classification in low labeled-data regimes.
    
    \item A dynamic training schedule that alternates between supervised phases and unsupervised phases to optimize learning efficiency.
    
    \item An image-to-image translation mechanism employed as a secondary task that, unlike purely generative approaches that create images de novo from noise vectors, modifies existing real unlabeled images to preserve authentic anatomical features while enriching feature representations beyond what traditional generative approaches provide.
    
    \item A confidence-weighted temporal ensemble technique that combines predictions from multiple model components and previous training iterations, significantly improving pseudo-labeling reliability in low-data scenarios.
    
    \item A comprehensive empirical evaluation demonstrating competitive performance against six state-of-the-art SSL methods across eleven benchmark datasets for medical image classification tasks.
\end{itemize}
The remainder of this paper is organized as follows: section \ref{sec:RL} reviews the related works in the field, providing context for our research contributions. Section \ref{sec:MT} details our proposed method. Section \ref{sec:EXC} describes the experimental configuration, including datasets, parameters, and evaluation metrics. Section \ref{sec:RS} presents our results and provides comprehensive analysis. Finally, section \ref{sec:CON} concludes the paper with a summary of our findings and suggestions for future research directions.

\section{Related Works}
\label{sec:RL}
Semi-supervised learning methods aim to leverage both labeled and unlabeled data to improve model performance, particularly in scenarios where labeled data is scarce or expensive to obtain.
A recent survey \cite{vanEngelen2020} has established a clear taxonomy of SSL approaches, distinguishing between two main classes: inductive methods, which construct classifiers that can generate predictions for any input, and transductive methods, which optimize directly over predictions for a given set of unlabeled data points.
Inductive methods can be further subdivided into three categories based on how they incorporate unlabeled data: (1) wrapper methods, which iteratively train classifiers on labeled data and use their predictions to generate pseudo-labels for unlabeled samples; (2) unsupervised preprocessing methods, which extract features or determine initial parameters from unlabeled data before supervised training; and (3) intrinsically semi-supervised methods, which directly incorporate unlabeled data into the objective function or optimization procedure.
Since our proposed approach falls within this third category, and it exploits GANs, the remainder of this section reviews the GAN-based SSL methods across various domains, focusing on approaches that have introduced key architectural innovations, while the interested readers can refer to~\cite{SSL_Surv} for a comprehensive review of approaches within this categories. 
Semi-supervised learning is built on the fundamental assumption that the data distribution in the input space contains substantial information about label distribution in the output space.
Within this context, GANs are particularly suitable candidates for SSL applications, given their inherent ability to model underlying data distributions and reveal patterns in the input space.
The first significant work in this context introduced the  SGAN model~\cite{SGAN}, which expands the traditional GAN architecture by augmenting the discriminator to perform dual functions that  distinguishes between real and synthetic samples while simultaneously predicting class labels for input data.
This dual-purpose approach represented an important extension of the framework through pseudo-labeling, where the discriminator/classifier is trained on both labeled data and generated samples with known class labels.
SGAN exemplifies what is known as a two-player model in GAN-based SSL, where the traditional generator-discriminator architecture is maintained but the discriminator is extended to perform both adversarial discrimination and classification tasks simultaneously.
Building on these foundations, MatchGAN \cite{MatchGAN} introduced an innovative approach that leveraged the Wasserstein distance and conditional generation.
As a semi-supervised conditional GAN, MatchGAN utilizes the label space in the target domain along with unlabeled samples to generate additional labeled training data.
The framework assigns labels from the pool of labeled samples to unlabeled samples, then passes these through the generator to create synthetic versions of images based on the target labels.
This work also introduces a match loss term that compares the generated images to the original labeled images from which the target labels are sampled.
A breakthrough in GAN-based SSL came with the introduction of TripleGAN \cite{TripleGAN}, which addressed the difficulty of simultaneously optimizing both generator and discriminator performance.
TripleGAN pioneered the three-player model architecture by incorporating an additional classifier that works independently from the discriminator, creating a tripartite interaction between generator, discriminator, and classifier.
In this three-player setup, the classifier works in conjunction with the generator to characterize conditional distributions between images while limiting the discriminator's role to identifying fake image-label pairs.
This separation of concerns allows each component to specialize in its primary task, potentially leading to improved overall performance compared to two-player models where the discriminator must balance competing objectives.
Recent developments have significantly expanded upon the TripleGAN framework~\cite{ECGAN, SECCGAN, CISSLGAN}.
EC-GAN \cite{ECGAN} proposed a mechanism where generated images are immediately processed by a classifier to produce pseudo-labels.
This classifier-generator interaction is regulated through a hyperparameter-weighted loss function that precisely controls the influence of generated samples on classifier training.
SEC-CGAN \cite{SECCGAN}  introduced a co-supervised learning paradigm where a conditional GAN is trained alongside the classifier, providing semantics-conditioned, confidence-aware synthesized examples during training.
CISSL-GAN \cite{CISSLGAN} then extended the Triple-GAN framework to address semi-supervised learning with class-imbalanced data through a dynamic class-rebalancing sampler that strategically selects pseudo-labeled samples from unlabeled data.
The analysis of the literature reported so far shows that the following   several  open issues still exist in current GAN-based SSL approaches:
The analysis of the literature reported so far shows that the following   several  open issues still exist in current GAN-based SSL approaches:
\begin{itemize}
    \item  None of the existing methods specifically addresses the challenges of extremely low labeled data regimes, such as the case where only  5-10 labeled samples per class are available.
    
    \item Current approaches like SGAN \cite{SGAN}, TripleGAN \cite{TripleGAN}, and EC-GAN \cite{ECGAN} primarily rely on generation-based paradigms for unsupervised learning, lacking effective mechanisms to integrate supervised and unsupervised signals. 
    
    \item Existing methods rely on a single discriminator or  on a single  classifier, while none has investigated possible advantages given by the use of an ensemble of models that, in other domain, has proven to provide complementary outputs that enhance model robustness~\cite{importance_ensemble,guarrasi2022pareto}.

\end{itemize}

Next section introduces our methodology that addresses these limitations.

\section{Methods} \label{sec:MT}
\begin{figure*}[t]
    \centering
    \includegraphics[width=\textwidth]{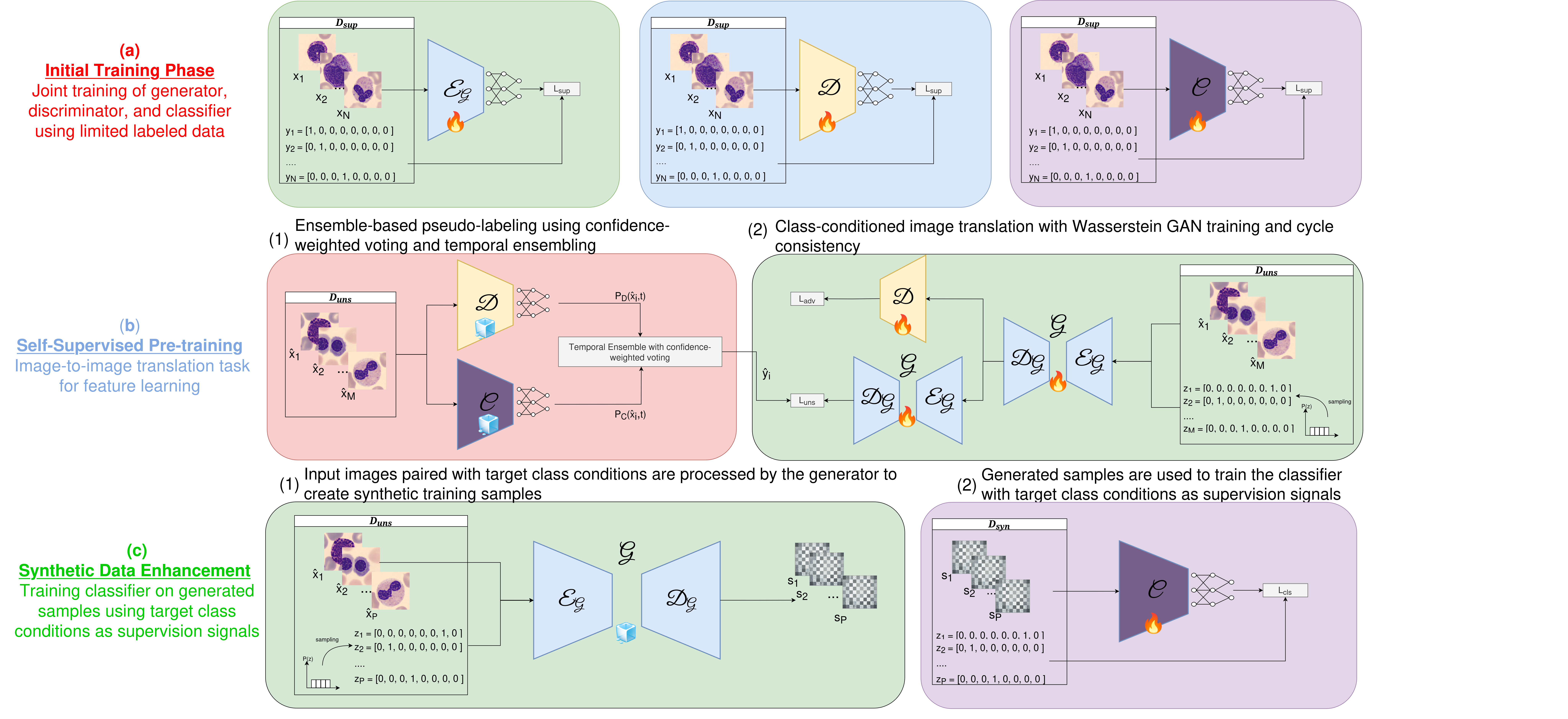}
    \caption{Three-phase framework for semi-supervised learning with limited labeled data. Our approach integrates three specialized networks: a generator ($\mathscr{G}$) for class-conditioned image synthesis, a discriminator ($\mathscr{D}$) for authenticity assessment and classification signalling, and a dedicated classifier ($\mathscr{C}$) for the primary classification task. The generator comprises an encoder ($\mathscr{E_{G}}$) and decoder ($\mathscr{D_{G}}$) for image-to-image translation.(a) Initial Training Phase: Joint training of these three networks using limited labeled data. (b) Self-Supervised Pre-training: Two-part approach combining (1) ensemble-based pseudo-labeling using confidence-weighted voting and temporal ensembling, and (2) class-conditioned image translation with Wasserstein GAN training and cycle consistency. (c) Synthetic Data Enhancement Phase: Training classifier on generated samples using target class conditions as supervision signals, where (1) input images paired with target class conditions are processed by the generator to create synthetic training samples, and (2) generated samples are used to train the classifier with target class conditions as supervision signals. Fire symbols indicate trainable networks while ice symbols represent frozen weights during respective training phases.}
    \label{fig:framework}
\end{figure*}

We propose a novel semi-supervised learning strategy called SPARSE (\textbf{S}emi-supervised \textbf{P}seudo-labeling via \textbf{A}dversarial \textbf{R}epresentation tran\textbf{S}lation \textbf{E}nhancement), designed to achieve robust classification performance in extremely low labeled-data regimes. 
Our approach integrates three specialized neural networks: a generator ($\mathscr{G}$) that performs class-conditioned image translation, a discriminator ($\mathscr{D}$) that assesses image authenticity while providing classification signals, and a dedicated classifier ($\mathscr{C}$) that focuses exclusively on the classification task.
Our approach consists of three main phases: 
\begin{enumerate}
    \item Supervised training phase (Figure~\ref{fig:framework}a): it jointly trains the  three aforementioned networks, $\mathscr{G}, \mathscr{D}$ and $\mathscr{C}$. 
    Each model is trained with a supervised dataset   $D_{sup} = \{(x_i, y_i)\}_{i=1}^{N}$, where $N$ is the number of few-shot samples, $x_i$ is the input sample - an image in our experiments, and  $y_i$ is the corresponding one-hot encoded ground truth vector, with $y_i \in \{0,1\}^{K \times 1}$ where $K$ is the total number of classes. 
    This initial phase is crucial for maintaining classification accuracy and preventing drift in the unsupervised learning process,  represented in panel b of the same figure,  by providing supervised signals from the limited labeled data.
    \item  Self-supervised pre-training phase (Figure~\ref{fig:framework}b), which  consists of two components.
    The first is an ensemble-based pseudo-labeling  block that combines the confidence-weighted scores provided by $\mathscr{D}$ and $\mathscr{C}$ given a set $D_{uns} = \{\hat{x}_i\}_{i=1}^{M}$ of $M$ unsupervised samples, with $M >> N$. 
    This ensemble outputs the pseudo-labels $\{\hat{y}_i\}_{i=1}^{M}$, with $\hat{y}_i \in \{0,1\}^{K \times 1}$,  assigned to each sample in $D_{uns}$.
    The second component of the self-supervised pretraining phase introduces a class-conditioned image translation task that uses randomly sampled class conditions $\{z_i\}_{i=1}^{M}$, with $z_i \in \{0,1\}^{K \times 1}$. 
    Hence, this phase leverages the abundant unlabeled data to improve feature representations and model generalization capabilities through image-to-image translation tasks.   
    \item Synthetic data enhancement phase (Figure~\ref{fig:framework}c): it receives as input $D_{uns}$ paired with one-hot encoded class vectors $\{z_i\}_{i=1}^{P}$ randomly sampled from a uniform distribution, which are processed by the generator $\mathscr{G}$ to create synthetic training samples $D_{syn} = \{s_i\}_{i=1}^{P}$, with $P>>N$.
    Subsequently, $D_{syn}$ is used to train the classifier $\mathscr{C}$ with these same class vectors $\{\hat{z}_i\}_{i=1}^{P}$ serving as supervision signals.
    This final phase aims to expand the effective training set by creating synthetic samples that augment the limited labeled data.
\end{enumerate}

The training schedule alternates between two phases: the supervised phase (executed at every epoch) and the combined self-supervised and synthetic data enhancement phase (executed every $\mu$ epochs, where $\mu$ is a hyperparameter), ensuring stable and effective utilization of the entire dataset.

It is worth noting that our approach addresses two primary issues in semi-supervised learning in an extremely low labeled-data regime.
The first  is the availability of insufficient labeled samples for effective supervised learning, which may affect panel (a) of Figure~\ref{fig:framework}: the self-supervised pretraining in panel (b) synthesizes new samples that are used in panel (c) to train the classifier with a large amount of samples.
The second issue concerns the integration of supervised and unsupervised learning signals.  While the paradigm  in the literature \cite{SGAN, TripleGAN, ECGAN} defines pre-tasks where a generative model  conditionally generates new samples that enhance the downstream classification task, 
our approach integrates an image-to-image translation pre-task that enrich the model with more  semantic information than a standard generative step (Figure~\ref{fig:framework}b). 

The rest of this section details such three phases: next subsection~\ref{subsec:supervisedTrainingPhase} presents panel (a) of Figure~\ref{fig:framework}, whilst subsection~\ref{subsec:UnsupervisedTrainingPhase} described both panels (b) and (c) of the same figure. 
Furthermore, subsection~\ref{subsec:inf_conf} introduces how to use our approach in  inference.

\subsection{Supervised Training Phase} \label{subsec:supervisedTrainingPhase}
The supervised phase in (Figure~\ref{fig:framework}a) is crucial for maintaining classification accuracy and preventing drift in the unsupervised learning process. 
It occurs at every epoch, utilizing the limited labeled data to train simultaneously the encoder of the generator $E_{\mathscr{G}}$, $\mathscr{D}$ and $\mathscr{C}$ to perform classification. 
We leverage deep supervision by adding a classification tail to the bottleneck of the generator's encoder $E_{\mathscr{G}}$ denoted as a set of interconnected neurons in panel (a) of Figure~\ref{fig:framework}; it is located at the bottleneck because it serves as the information compression point between the encoder and decoder paths.
By applying classification supervision at this critical juncture, we ensure that the most compact representation in the network encodes both structural information needed for generation and semantic information required for classification. \\

The supervised loss function use by all the three networks ($\mathcal{L}_{sup}$) combines four specialized loss terms, each addressing a specific challenge in few-shot learning and weighted by coefficients to balance their contribution. 
\begin{equation}
\label{eqn:l_sup}
\mathcal{L}_{sup} = \mathcal{L}_{prototype} + \alpha\mathcal{L}_{mutual} + \beta\mathcal{L}_{entropy} + \gamma\mathcal{L}_{mixup}
\end{equation}

The prototype loss ($\mathcal{L}_{prototype}$) helps create discriminative class-specific features by learning robust prototypical representations for each class. 
The mutual learning loss ($\mathcal{L}_{mutual}$) enables knowledge sharing between the three models, leveraging their complementary perspectives on the data. The entropy minimization loss ($\mathcal{L}_{entropy}$) encourages the models to make confident predictions, helping combat the uncertainty inherent in limited-data scenarios. 
Finally, the mixup loss ($\mathcal{L}_{mixup}$) provides regularization through data augmentation, helping prevent overfitting which is particularly crucial when training with few samples. 
The rest of this section details the computation of each of these four losses functions. \\

The prototype loss $\mathcal{L}_{prototype}$ creates discriminative class-specific representations by comparing softmax probabilities with class prototypes:
\begin{equation}
\mathcal{L}_{prototype} = -\sum_{i=1}^N \log\frac{\exp(-d(p_T(x_i), c_{y_i}))}{\sum_{k=1}^K \exp(-d(p_T(x_i), c_k))}
\end{equation}
where $p_T(x_i)$ represents the softmax probabilities of input $x_i$ with temperature $T$, $c_k$ is the prototype of class $k$ computed as the mean of the class probabilities:
\begin{equation}
c_k = \frac{1}{|S_k|}\sum_{x_i \in S_k} p_T(x_i)
\end{equation}
where $S_k$ is the set of samples from class $k$. The distance function $d(·,·)$ is defined as the negative sum of squared differences:
\begin{equation}
d(p_T(x), c_k) = -\sum_{j=1}^K (p_{T,j}(x) - c_{k,j})^2
\end{equation}
where $K$ is the total number of classes. \\

The mutual learning loss $\mathcal{L}_{mutual}$ facilitates knowledge transfer between models through a combination of supervised cross-entropy and KL divergence:
\begin{equation}
\mathcal{L}_{mutual} = \mathcal{L}_{ce} + \lambda_{kl}\mathcal{L}_{kl}
\end{equation}
where $\lambda_{kl}$ is the weight coefficient for the KL divergence term, $\mathcal{L}_{ce}$ is the sum of cross-entropy losses for each model:
\begin{equation}
\mathcal{L}_{ce} = \sum_{m \in \{E_{\mathscr{G}},\mathscr{D},\mathscr{C}\}} \text{CE}(p_m(x), y)
\end{equation}
and $\mathcal{L}_{kl}$ is the symmetric KL divergence between each model's probabilities and the average of other models' probabilities:
\begin{equation}
\mathcal{L}_{kl} = \sum_{m \in \{E_{\mathscr{G}},\mathscr{D},\mathscr{C}\}} \text{KL}(p_m(x) | \frac{1}{2}\sum_{n \neq m} p_n(x))
\end{equation} \\

The entropy minimization loss $\mathcal{L}_{entropy}$ promotes confident predictions:
\begin{equation}
\mathcal{L}_{entropy} = -\sum_{k=1}^K p_k(x)\log(p_k(x))
\end{equation} \\

Finally, the mixup loss $\mathcal{L}_{mixup}$ provides regularization by training on interpolated samples and labels. 
For each pair of samples $(x_i,y_i)$ and $(x_j,y_j)$, the interpolation weight $\lambda_{mix}$ is sampled from a Beta distribution:
\begin{equation}
\lambda_{mix} \sim \text{Beta}(\alpha_{mix}, \alpha_{mix})
\end{equation}

The Beta distribution is particularly suitable for generating interpolation weights as it is bounded between [0,1] and can be symmetric around 0.5, ensuring a balanced mixing of samples while maintaining their relative contributions. 
The hyperparameter $\alpha_{mix}$ in the Beta distribution controls the strength of interpolation - higher values of $\alpha_{mix}$ lead to interpolation weights closer to 0.5, while lower values favour weights closer to 0 or 1.
This weight is then used to create interpolated samples and labels:
\begin{equation}
\tilde{x} = \lambda_{mix} x_i + (1-\lambda_{mix})x_j
\end{equation}
\begin{equation}
\tilde{y} = \lambda_{mix} y_i + (1-\lambda_{mix})y_j
\end{equation}

The mixup loss is then computed as:
\begin{equation}
\mathcal{L}_{mixup} = \mathcal{L}_{ce}(p(\tilde{x}), \tilde{y})
\end{equation}
where $\mathcal{L}_{ce}$ is the cross-entropy loss. 

\subsection{Unsupervised Training Phase}
\label{subsec:UnsupervisedTrainingPhase}
The unsupervised training phase, illustrated in panel (b) of Figure~\ref{fig:framework}, executes every $\mu$ epochs. This phase leverages unlabeled data through an image-to-image translation framework with three distinct stages. 

First, we use an ensemble-based pseudo-labeling mechanism to estimate class probabilities for unlabeled samples (subsection \ref{subsubsec:Ensemble}). These probability estimates then guide a class-conditioned image translation process, which learns to generate class-specific variations of input images (subsection \ref{subsubsec:Class-Conditioned-Image-Translation}). Finally, we employ these generated samples in a synthetic data enhancement phase. Here, the synthetic images serve as additional training data to strengthen the classifier's ability to distinguish between classes (subsection \ref{subsubsec:Synthetic-Data-Ehnancement}).

\subsubsection{Ensemble-based Pseudo-labeling}
\label{subsubsec:Ensemble}
To effectively utilize unlabeled samples, we require reliable class probability estimates. Our approach addresses three key challenges: (1) quantifying model uncertainty, (2) aggregating predictions from multiple models, and (3) maintaining temporal stability throughout training. We tackle these challenges through an ensemble mechanism (Figure~\ref{fig:framework}b) that integrates confidence-weighted voting, temporal ensembling, and adaptive thresholding.

At epoch $t$, each unlabeled image $\hat{x}_i$ from $D_{uns} = \{\hat{x}_i\}_{i=1}^{M}$ is processed by the models trained during the previous initial training phase. We apply temperature scaling to obtain calibrated class probabilities:
\begin{equation}
p_m(\hat{x}_i,t) = \text{softmax}\left(\frac{l_m(\hat{x}_i,t)}{T}\right)
\end{equation}
where $l_m(\hat{x}_i,t)$ represents the \textbf{logits} (raw pre-softmax outputs) from model $m \in \{\mathscr{D}, \mathscr{C}\}$ for input $\hat{x}_i$ at epoch $t$. The temperature parameter $T$ controls prediction sharpness: lower values ($T < 1$) yield more confident predictions, while higher values ($T > 1$) produce smoother probability distributions.

Each model's prediction reliability is quantified using an entropy-based confidence measure:
\begin{equation}
c_m(\hat{x}_i,t) = 1 - \frac{H(p_m(\hat{x}_i,t))}{H_{max}}
\end{equation}
where the entropy $H(p_m(\hat{x}_i,t))$ is computed as:
\begin{equation}
H(p_m(\hat{x}_i,t)) = -\sum_{k=1}^{K} p_m^{(k)}(\hat{x}_i,t) \log p_m^{(k)}(\hat{x}_i,t)
\end{equation}
and $H_{max} = \log K$ represents the maximum possible entropy for $K$ classes. The term $p_m^{(k)}(\hat{x}_i,t)$ denotes model $m$'s predicted probability for class $k$. This confidence score ranges from 0 (complete uncertainty) to 1 (complete certainty).

We combine predictions from both models by weighting each according to its confidence score:
\begin{equation}
p_{weighted}(\hat{x}_i,t) = \frac{\sum_{m \in \{\mathscr{D},\mathscr{C}\}} c_m(\hat{x}_i,t) \cdot p_m(\hat{x}_i,t)}{\sum_{m \in \{\mathscr{D},\mathscr{C}\}} c_m(\hat{x}_i,t)}
\end{equation}
The denominator ensures proper normalization. This weighting scheme allows more confident models to contribute more strongly to the final prediction.

To enhance prediction stability, we incorporate historical information by blending current predictions with past predictions:
\begin{equation}
p_{ens}(\hat{x}_i,t) = \alpha \cdot p_{weighted}(\hat{x}_i,t) + (1-\alpha) \cdot p_{ema}(\hat{x}_i,t-1)
\end{equation}
Here, $p_{ema}(\hat{x}_i,t-1)$ captures the temporal history through an exponential moving average (EMA) of past ensemble predictions. The parameter $\alpha \in [0,1]$ balances current information (higher $\alpha$) against historical stability (lower $\alpha$). This temporal smoothing prevents abrupt prediction changes that could destabilize the training process.

After computing the current ensemble prediction, we update the EMA:
\begin{equation}
p_{ema}(\hat{x}_i,t) = \beta \cdot p_{ema}(\hat{x}_i,t-1) + (1-\beta) \cdot p_{ens}(\hat{x}_i,t)
\end{equation}
The momentum parameter $\beta \in [0,1]$ determines the temporal memory span. Large values (e.g., $\beta = 0.99$) maintain longer memory for stable predictions, while smaller values allow faster adaptation to recent changes.

With reliable probability estimates in hand, we select only the most confident predictions for pseudo-labeling:
\begin{equation}
\mathcal{S}(t) = \{\hat{x}_i \in D_{uns} : \max_k(p_{ens}^{(k)}(\hat{x}_i,t)) > \tau(t)\}
\end{equation}
where $\max_k(p_{ens}^{(k)}(\hat{x}_i,t))$ is the highest class probability from the ensemble prediction. The subset $\mathcal{S}(t)$ contains selected samples at epoch $t$, and the adaptive threshold $\tau(t)$ is defined as:
\begin{equation}
\tau(t) = Q_{\rho}(\{\max_k(p_{ens}^{(k)}(\hat{x}_i,t)) : \hat{x}_i \in D_{uns}\})
\end{equation}
Here, $Q_{\rho}$ denotes the $\rho$-th percentile of maximum probabilities across all unlabeled samples in the current batch, where $\rho \in [0,1]$ is the percentile threshold parameter.

This percentile-based approach automatically adjusts to the model's current performance level. It selects $(1-\rho) \times 100\%$ of the most confident samples. For instance, setting $\rho = 0.8$ selects the top 20\% most confident predictions. This adaptive mechanism prevents error accumulation from unreliable pseudo-labels while naturally accommodating the model's improving performance.

For each selected confident sample $\hat{x}_i \in \mathcal{S}(t)$, we create a discrete pseudo-label:
\begin{equation}
\hat{y}_i = \text{one-hot}(\arg\max_k p_{ens}^{(k)}(\hat{x}_i,t))
\end{equation}
This produces a one-hot encoded pseudo-label $\hat{y}_i \in \{0,1\}^{K \times 1}$ for each unlabeled sample $\hat{x}_i$. The complete set of pseudo-labels $\{\hat{y}_i\}_{i=1}^{|\mathcal{S}(t)|}$ for all selected samples then feeds into the subsequent class-conditioned image translation process.

\subsubsection{Class-conditioned Image Translation}
\label{subsubsec:Class-Conditioned-Image-Translation}
Using the pseudo-labels $\{\hat{y}_i\}_{i=1}^{M}$ obtained from our ensemble mechanism, we implement a class-conditioned image translation process that leverages $\mathscr{G}$ and $\mathscr{D}$ working in tandem (Figure~\ref{fig:framework}b right).
Now the generator $\mathscr{G}$, which consists of a complete U-Net architecture, not just the encoder as in the supervised phase, learns to perform class-conditioned image translation while preserving semantic features relevant to classification.
Its training objective $\mathcal{L}_{uns}$ combines four components:
\begin{equation}
\mathcal{L}_{uns} = \mathcal{L}_{adv} + \mathcal{L}_{cls}^{\mathscr{D}} + \mathcal{L}_{cls}^{\mathscr{G}} + \lambda_{rec}\mathcal{L}_{rec}
\end{equation}
where $\lambda_{rec}$ is the weight coefficient for the reconstruction loss, and the adversarial loss $\mathcal{L}_{adv}$ uses the Wasserstein distance metric to assess image realism:
\begin{equation}
\mathcal{L}_{adv} = -\mathbb{E}_{\hat{x} \sim D_{uns}}[\mathscr{D}(\mathscr{G}(\hat{x}, z_{target}))],
\end{equation}

Next, the classification losses $\mathcal{L}_{cls}^{\mathscr{D}}$ and $\mathcal{L}_{cls}^{\mathscr{G}}$ ensure accurate conditioning on target classes using cross-entropy from both the discriminator and generator classifiers:
\begin{equation}
\mathcal{L}_{cls}^{\mathscr{D}} = -\mathbb{E}_{\hat{x} \sim D_{uns}}\sum_{k=1}^K z_{target}^k \log(p_k^{\mathscr{D}}(\mathscr{G}(\hat{x}, z_{target})))
\end{equation}
\begin{equation}
\mathcal{L}_{cls}^{\mathscr{G}} = -\mathbb{E}_{\hat{x} \sim D_{uns}}\sum_{k=1}^K z_{target}^k \log(p_k^{\mathscr{G}}(\mathscr{G}(\hat{x}, z_{target})))
\end{equation}

where $z_{target}^k$ is the $k_{th}$ element of the target class one-hot encoding, $p_k^{\mathscr{D}}$ represents the probability for class \textit{k} from the discriminator, and $p_k^{\mathscr{G}}$ represents the probability for class \textit{k} from the generator's classifier.
The reconstruction loss $\mathcal{L}_{rec}$ maintains content consistency using L1 distance:
\begin{equation}
\mathcal{L}_{rec} = \mathbb{E}_{\hat{x} \sim D_{uns}}[|\mathscr{G}(\mathscr{G}(\hat{x}, z_{target}), z_{source}) - \hat{x}|_1]
\end{equation}

where $z_{source}$ is the one-hot encoding of the most probable class according to $p_{ens}(\hat{x},t)$, connecting this translation process with our previous pseudo-labeling step. \\

The discriminator $\mathscr{D}$ serves a dual role: it assesses image realism through a Wasserstein distance metric while also providing classification signals. 
Its objective function reflects these dual tasks:
\begin{equation}
\mathcal{L}_{\mathscr{D}} = -\mathcal{L}_{adv} + \lambda_{cls}\mathcal{L}_{cls} + \lambda_{gp}\mathcal{L}_{gp}
\end{equation}

where $\lambda_{cls}$ and $\lambda_{gp}$ are weight coefficients for the classification loss and gradient penalty terms, respectively. Its adversarial component implements the Wasserstein distance as follows:
\begin{equation}
-\mathcal{L}_{adv} = -\mathbb{E}_{\hat{x} \sim D_{uns}}[\mathscr{D}(\hat{x})] + \mathbb{E}_{\hat{x} \sim D_{uns}}[\mathscr{D}(\mathscr{G}(\hat{x}, z_{target}))]
\end{equation}

The gradient penalty term $\mathcal{L}_{gp}$ enforces the Lipschitz constraint:
\begin{equation}
\mathcal{L}_{gp} = \mathbb{E}_{\hat{x}_{interp}}[(|\nabla_{\hat{x}_{interp}}\mathscr{D}(\hat{x}_{interp})|_2 - 1)^2]
\end{equation}

where $\hat{x}_{interp}$ is sampled uniformly along straight lines between pairs of real and generated images. \\

\subsubsection{Synthetic Data Enhancement}
\label{subsubsec:Synthetic-Data-Ehnancement}
After establishing the image translation process, we leverage translated images as a form of synthetic labeled data to enhance the classifier's performance (Figure~\ref{fig:framework}c). 
The key process involves taking unlabeled input images $\{\hat{x}_i\}_{i=1}^{M}$ and sampling random target classes  $\{z_{i}\}_{i=1}^{P}$ from a uniform distribution over the $K$ classes in one-hot format, where $P$ is the number of synthetic samples generated. 
We then use these randomly sampled classes to condition the generator to translate the original images of unknown classes.
The resulting translated images have known target classes $\{z_{i}\}_{i=1}^{P}$. 

These translated images with known target classes can then be used to train the classifier in a supervised manner:
\begin{equation}
\mathcal{L}_{\mathscr{C}} = \mathcal{L}_{cls}(\mathscr{C}(\mathscr{G}(\hat{x}, z)), z)
\end{equation}

where $z$ represents the target class condition provided to the generator for synthetic data generation.
This approach provides us with effectively labeled training samples, since we know exactly what class condition was used to generate each image.
Importantly, as demonstrated in \cite{GoodBadGAN}, the visual quality of generated images is not critical for our classification objective: 
instead, we focus on ensuring that the translation process captures and preserves discriminative features that are useful for classification.

\subsection{Inference configuration}
\label{subsec:inf_conf}
Our approach in the inference phase can be deployed in different configurations that exploits the model trained as reported in section ~\ref{sec:MT}.
In particular, in the rest of the manuscript we consider the following two set-ups for inference:
\begin{itemize}
    \item SPARSE: it used only the classifier $\mathscr{C}$.
    \item SPARSE\textsubscript{ens}: it exploits both the discriminator $\mathscr{D}$ and the classifier $\mathscr{C}$, which are combined in late fusion by averaging the estimates of posterior probabilities per class.
\end{itemize}

It is worth noting that in our ensemble configuration, we should also consider the potential use of the generator's encoder  $\mathscr{E_G}$.
However, since it is trained to minimizer a loss function that balances both generation and classification objectives, this results in a performance degradation - a finding we experimentally verified, though omitted from the manuscript for conciseness.

\begin{table*}[ht]
\centering
\caption{Characteristics of MedMNIST datasets used in our experiments.}
\label{tab:datasets}
\resizebox{\textwidth}{!}{
\begin{tabular}{@{}l@{\hspace{1em}}l@{\hspace{1em}}l@{\hspace{1.5em}}r@{\hspace{1em}}r@{\hspace{1em}}r@{}}
\toprule
\multicolumn{1}{c}{\multirow{2}{*}{\textbf{Dataset}}} & \multicolumn{1}{c}{\multirow{2}{*}{\textbf{Modality}}} & \multicolumn{1}{c}{\multirow{2}{*}{\textbf{Task Type}}} & \multicolumn{3}{c}{\textbf{\# Samples}} \\
\cmidrule(lr){4-6}
 &  &  & \multicolumn{1}{c}{\textbf{Total}} & \multicolumn{1}{c}{\textbf{Training}} & \multicolumn{1}{c}{\textbf{Val/Test}} \\
\midrule
BloodMNIST & Blood Cell Microscope & Multi-Class (8) & 17,092 & 11,959 & 1,712/3,421 \\[0.5ex]
BreastMNIST & Breast Ultrasound & Binary-Class (2) & 780 & 546 & 78/156 \\[0.5ex]
ChestMNIST & Chest X-Ray & Binary-Class (2) & 112,120 & 78,468 & 11,219/22,433 \\[0.5ex]
DermaMNIST & Dermatoscope & Multi-Class (7) & 10,015 & 7,007 & 1,003/2,005 \\[0.5ex]
OCTMNIST & Retinal OCT & Multi-Class (4) & 109,309 & 97,477 & 10,832/1,000 \\
\addlinespace[0.5ex]
OrganAMNIST & Abdominal CT & Multi-Class (11) & 58,830 & 34,561 & 6,491/17,778 \\[0.5ex]
OrganCMNIST & Abdominal CT & Multi-Class (11) & 23,583 & 12,975 & 2,392/8,216 \\[0.5ex]
OrganSMNIST & Abdominal CT & Multi-Class (11) & 25,211 & 13,932 & 2,452/8,827 \\
\addlinespace[0.5ex]
PathMNIST & Colon Pathology & Multi-Class (9) & 107,180 & 89,996 & 10,004/7,180 \\[0.5ex]
PneumoniaMNIST & Chest X-Ray & Binary-Class (2) & 5,856 & 4,708 & 524/624 \\[0.5ex]
TissueMNIST & Kidney Cortex Microscope & Multi-Class (8) & 236,386 & 165,466 & 23,640/47,280 \\
\bottomrule
\end{tabular}
}
\end{table*}

\section{Experimental Configuration}
\label{sec:EXC}
This section describes our experimental methodology: it details the materials (subsection \ref{subsec:mat}), followed by our training configuration (subsection \ref{subsec:tr_conf}).

\subsection{Materials}
\label{subsec:mat}
We conducted experiments using eleven datasets from MedMNIST repository \cite{medmnist}: BloodMNIST, BreastMNIST, ChestMNIST, DermaMNIST, OCTMNIST, OrganAMNIST, OrganCMNIST, OrganSMNIST, PathMNIST, PneumoniaMNIST and TissueMNIST.
As shown in Table \ref{tab:datasets}, these  datasets span different medical imaging modalities and classification tasks, with varying number of classes (2-11) and dataset sizes (from hundreds to hundreds of thousands of samples). \\
To evaluate our method's effectiveness in extremely low labeled-data regimes, we conducted experiments across four few-shot settings: 5-shot, 10-shot, 20-shot, and 50-shot per class, with 5-shot representing the most challenging scenario.
For each \textit{N}-shot setting, we constructed the training set using \textit{N} labeled samples per class, with the remaining samples treated as unlabeled data.
For data preprocessing, we applied a transformation pipeline consisting of random horizontal flipping for data augmentation and tensor conversion, with input images maintaining their original 128×128 pixel resolution. 
To ensure reproducibility, we utilized the original validation/test splits provided by the MedMNIST authors.

\subsection{Training Configuration}
\label{subsec:tr_conf}
The training schedule alternates between supervised and unsupervised learning phases. The supervised phase occurs at every epoch, while the unsupervised phase is executed every $\mu$ epochs as already described in section \ref{sec:MT}. 
All models were trained for 1000 epochs using the AdamW optimizer, , maintaining identical configurations
throughout all experiments which are reported in table \ref{tab:hyperparameters}. 
During training, we implemented a model checkpoint strategy that saved the model state achieving the best validation accuracy, which was then used for final evaluation. 
For all the models, we did not investigate any hyperparameter configuration since their tuning is out of the scope of this manuscript.
Nevertheless, because the 'No Free Lunch' Theorem for optimization \cite{NFL} states that no universal set of hyperparameters will optimize a model's performance across all possible datasets, this approach ensures a fair comparison among all the approaches.
\begin{table}[ht]
\centering
\caption{Training hyperparameters used in our experiments}
\label{tab:hyperparameters}
\begin{tabular}{lc}
\hline
\textbf{Parameter} & \textbf{Value} \\
\hline
\multicolumn{2}{c}{\textit{General Training Parameters}} \\
\hline
Training Epochs & 1000 \\
Base Learning Rate & 0.0002 \\
Optimizer & AdamW \\
$\mu$ (unsupervised phase frequency) & 10 \\
$T$ (temperature parameter) & 2.0 \\
\hline
\multicolumn{2}{c}{\textit{Loss Weights for Supervised Objective}} \\
\hline
$\alpha$ (mutual learning loss weight) & 0.1 \\
$\beta$ (entropy minimization loss weight) & 0.01 \\
$\gamma$ (mixup loss weight) & 0.5 \\
$\lambda_{kl}$ (KL divergence weight) & 0.5 \\
\hline
\multicolumn{2}{c}{\textit{mixup parameters}} \\
\hline
$\alpha_{mix}$ (beta distribution parameter) & 0.2 \\
\hline
\multicolumn{2}{c}{\textit{loss weights for generator}} \\
\hline
$\lambda_{rec}$ (reconstruction loss weight) & 10.0 \\
\hline
\multicolumn{2}{c}{\textit{loss weights for discriminator}} \\
\hline
$\lambda_{cls}$ (classification loss weight) & 1.0 \\
$\lambda_{gp}$ (gradient penalty weight) & 10.0 \\
\hline
\multicolumn{2}{c}{\textit{Ensemble Parameters}} \\
\hline
$\alpha$ (temporal ensemble weight) & 0.6 \\
$\beta$ (EMA momentum) & 0.99 \\
$\rho$ (percentile threshold) & 0.75 \\
\hline
\end{tabular}
\end{table}

\section{Results}
\label{sec:RS}
This section presents the experimental results evaluating our framework's effectiveness in extremely low labeled data regimes in medical image classification.
We compare our approach against the six state-of-the-art semi-supervised GAN architectures which are already presented in section~\ref{sec:RL}, using their original implementations without additional optimization or modifications to ensure fair comparison with our method.

To assess model performance, we employ classification accuracy per class as our primary evaluation metric. 
For each dataset, we first compute the accuracy across all available classes and then the final score is obtained by averaging these accuracies across all  datasets, providing a robust evaluation of the model's generalization capabilities across varying medical imaging tasks, modalities, and data distributions.
Individual results for each dataset are presented in the appendix.

Table~\ref{tab:performance_comparison} presents the performance of all models across different few-shot settings, where we systematically vary the number of samples per class from extremely limited (5-shot) to more moderate (50-shot) situations. 
The first two rows show the results of our approach using only the classifier $\mathscr{C}$ (SPARSE) or using the ensemble between the discriminator $\mathscr{D}$ and $\mathscr{C}$ (SPARSE\textsubscript{ens}).
The following six rows displays the results of the six competitors. 
We note that the two variants of our approach consistently outperform existing competitors across all settings, and the ensemble setting (SPARSE\textsubscript{ens}) achieves the best performance in all scenarios. 
As the number of labeled samples increases from 5 to 50 shots per class, we observe a steady improvement in performance across all models, though the relative advantage of our approach remains consistent.

\begin{table}[htbp]
    \centering
    \caption{Model Performance Across Different Few-Shot Settings}
    \label{tab:performance_comparison}
    \begin{tabular}{l|cccc}
    \toprule
    \multirow{2}{*}{Model} & \multicolumn{4}{c}{Average Accuracy per class} \\
    \cmidrule{2-5}
    & 5-shot & 10-shot & 20-shot & 50-shot \\
    \midrule
    SPARSE & 63.21 & 68.50 & 73.44 & 77.15 \\
    SPARSE\textsubscript{ens} & \textbf{66.22} & \textbf{70.95} & \textbf{75.71} & \textbf{78.28} \\
    \midrule
    SGAN \cite{SGAN} & 25.80 & 25.96 & 24.92 & 27.28 \\
    MatchGAN \cite{MatchGAN} & 39.88 & 48.73 & 51.90 & 54.15 \\
    EC-GAN \cite{ECGAN} & 35.09 & 34.66 & 34.29 & 47.41 \\
    TripleGAN \cite{TripleGAN} & 64.23 & 68.79 & 71.40 & 76.25 \\
    SEC-GAN \cite{SECCGAN} & 58.73 & 63.79 & 66.95 & 73.30 \\
    CISSL \cite{CISSLGAN} & 45.84 & 46.80 & 49.19 & 51.20 \\
    \bottomrule
    \end{tabular}
\end{table}

To further investigate these findings, the rest of this section examines our improvements in the 5-shot setting (section \ref{subsec:5shot}), 
which represents the most challenging scenario where the extreme scarcity of labeled data tests the true effectiveness of semi-supervised learning. 
We then complement the results with a detailed analysis in the 50-shot setting (section \ref{subsec:50shot}), which helps us understand how our method scales when more labeled data becomes available. 
Statistical significance testing of the performance differences is presented within these analyses.
Finally, section \ref{subsec:mu} investigates how the frequency of unsupervised training affects model performance.

\subsection{Performance in 5-shot}
\label{subsec:5shot}
We now deepen our analysis on the most challenging 5-shot learning scenario, where only 5 labeled samples per class are available for training while the remaining samples are treated as unlabeled. 
To provide robust statistical evaluation, we employ the Wilcoxon signed-rank test, a non-parametric method that compares paired accuracy values across our 11 datasets, with Benjamini-Hochberg FDR correction for multiple comparisons.
Table~\ref{tab:5shot_comparison} presents the results of this statistical comparison, with the upper triangular part showing corrected $p$-values with effect sizes ($r$) and their interpretations, where significant results ($p < 0.05$) are highlighted in bold, and the lower triangular part displaying Win-Tie-Loss (W-T-L) statistics from the row model's perspective when compared against the column model.

\begin{table*}[htbp]
    \centering
    \caption{5-Shot statistical comparison of model performance. The upper triangular part shows p-values from statistical comparison (significant values $p < 0.05$ highlighted in bold), while the lower triangular part shows Win-Tie-Loss (W-T-L) statistics. The r-value represents the effect size (Pearson's correlation coefficient) with interpretations: small ($r < 0.3$), medium ($0.3 \leq r < 0.5$), large ($0.5 \leq r < 0.7$), and very large ($r \geq 0.7$). Best performing model is highlighted in bold.}
    \label{tab:5shot_comparison}
    \footnotesize
    \renewcommand{\arraystretch}{1.1}
    \setlength{\tabcolsep}{3.5pt}
    \begin{tabular}{lcccccccc}
    \toprule
    \multirow{2}{*}{Model} & \multicolumn{8}{c}{Statistical Comparison (5 shot)} \\
    \cmidrule(lr){2-9}
    & SPARSE & \textbf{SPARSE\textsubscript{ens}} & SGAN & MatchGAN & CISSL & SEC-GAN & TripleGAN & ECGAN \\
    \midrule
    \multirow{3}{*}{SPARSE} & \multirow{3}{*}{-} & \multirow{3}{*}{\shortstack{\textbf{4.1e-02}\\r=0.818\\(very large)}} & \multirow{3}{*}{\shortstack{\textbf{2.1e-02}\\r=0.818\\(very large)}} & \multirow{3}{*}{\shortstack{6.0e-02\\r=0.636\\(very large)}} & \multirow{3}{*}{\shortstack{8.7e-02\\r=0.455\\(large)}} & \multirow{3}{*}{\shortstack{2.4e-01\\r=0.455\\(large)}} & \multirow{3}{*}{\shortstack{7.7e-01\\r=0.091\\(small)}} & \multirow{3}{*}{\shortstack{\textbf{4.4e-02}\\r=0.636\\(very large)}} \\
    \\
    \\
    \multirow{3}{*}{\textbf{SPARSE\textsubscript{ens}}} & \multirow{3}{*}{\shortstack{(10-0-1)}} & \multirow{3}{*}{-} & \multirow{3}{*}{\shortstack{\textbf{2.1e-02}\\r=1.000\\(very large)}} & \multirow{3}{*}{\shortstack{\textbf{4.4e-02}\\r=0.636\\(very large)}} & \multirow{3}{*}{\shortstack{\textbf{2.2e-02}\\r=0.818\\(very large)}} & \multirow{3}{*}{\shortstack{5.3e-02\\r=0.636\\(very large)}} & \multirow{3}{*}{\shortstack{1.5e-01\\r=0.455\\(large)}} & \multirow{3}{*}{\shortstack{\textbf{3.5e-02}\\r=0.800\\(very large)}} \\
    \\
    \\
    \multirow{3}{*}{SGAN} & \multirow{3}{*}{\shortstack{(1-0-10)}} & \multirow{3}{*}{\shortstack{(0-1-10)}} & \multirow{3}{*}{-} & \multirow{3}{*}{\shortstack{\textbf{2.2e-02}\\r=0.636\\(very large)}} & \multirow{3}{*}{\shortstack{\textbf{2.2e-02}\\r=0.636\\(very large)}} & \multirow{3}{*}{\shortstack{\textbf{2.1e-02}\\r=0.818\\(very large)}} & \multirow{3}{*}{\shortstack{\textbf{2.1e-02}\\r=0.818\\(very large)}} & \multirow{3}{*}{\shortstack{\textbf{4.4e-02}\\r=0.600\\(very large)}} \\
    \\
    \\
    \multirow{3}{*}{MatchGAN} & \multirow{3}{*}{\shortstack{(2-0-9)}} & \multirow{3}{*}{\shortstack{(2-0-9)}} & \multirow{3}{*}{\shortstack{(9-0-2)}} & \multirow{3}{*}{-} & \multirow{3}{*}{\shortstack{2.6e-01\\r=0.273\\(medium)}} & \multirow{3}{*}{\shortstack{6.0e-02\\r=0.455\\(large)}} & \multirow{3}{*}{\shortstack{6.0e-02\\r=0.455\\(large)}} & \multirow{3}{*}{\shortstack{1.0e-01\\r=0.636\\(very large)}} \\
    \\
    \\
    \multirow{3}{*}{CISSL} & \multirow{3}{*}{\shortstack{(3-0-8)}} & \multirow{3}{*}{\shortstack{(1-0-10)}} & \multirow{3}{*}{\shortstack{(9-0-2)}} & \multirow{3}{*}{\shortstack{(7-0-4)}} & \multirow{3}{*}{-} & \multirow{3}{*}{\shortstack{\textbf{4.1e-02}\\r=0.636\\(very large)}} & \multirow{3}{*}{\shortstack{\textbf{4.4e-02}\\r=0.455\\(large)}} & \multirow{3}{*}{\shortstack{\textbf{4.4e-02}\\r=0.455\\(large)}} \\
    \\
    \\
    \multirow{3}{*}{SEC-GAN} & \multirow{3}{*}{\shortstack{(3-0-8)}} & \multirow{3}{*}{\shortstack{(2-0-9)}} & \multirow{3}{*}{\shortstack{(10-0-1)}} & \multirow{3}{*}{\shortstack{(8-0-3)}} & \multirow{3}{*}{\shortstack{(9-0-2)}} & \multirow{3}{*}{-} & \multirow{3}{*}{\shortstack{8.7e-02\\r=0.455\\(large)}} & \multirow{3}{*}{\shortstack{\textbf{2.7e-02}\\r=0.818\\(very large)}} \\
    \\
    \\
    \multirow{3}{*}{TripleGAN} & \multirow{3}{*}{\shortstack{(5-0-6)}} & \multirow{3}{*}{\shortstack{(3-0-8)}} & \multirow{3}{*}{\shortstack{(10-0-1)}} & \multirow{3}{*}{\shortstack{(8-0-3)}} & \multirow{3}{*}{\shortstack{(8-0-3)}} & \multirow{3}{*}{\shortstack{(8-0-3)}} & \multirow{3}{*}{-} & \multirow{3}{*}{\shortstack{\textbf{4.4e-02}\\r=0.455\\(large)}} \\
    \\
    \\
    \multirow{3}{*}{ECGAN} & \multirow{3}{*}{\shortstack{(2-0-9)}} & \multirow{3}{*}{\shortstack{(1-1-9)}} & \multirow{3}{*}{\shortstack{(8-1-2)}} & \multirow{3}{*}{\shortstack{(2-0-9)}} & \multirow{3}{*}{\shortstack{(3-0-8)}} & \multirow{3}{*}{\shortstack{(1-0-10)}} & \multirow{3}{*}{\shortstack{(3-0-8)}} & \multirow{3}{*}{-} \\
    \\
    \\
    \bottomrule
    \end{tabular}
\end{table*}

The primary finding is that our ensemble model, SPARSE\textsubscript{ens}, achieves statistically significant superiority over most competing approaches, also with a number of wins larger than losses, demonstrating robust performance across diverse medical imaging datasets. \\
We now discuss the effectiveness of the ensemble approach by comparing SPARSE\textsubscript{ens} to its base model, SPARSE.
The results show that the ensemble model demonstrates a statistically significant improvement ($p=4.1 \times 10^{-2},r=0.818$), securing wins on 10 out of 11 datasets with only a single loss. 
This marked improvement is achieved through ensemble averaging at inference time. 
Although both configurations share an identical training procedure, the classifier $\mathscr{C}$ and discriminator $\mathscr{D}$ networks develop complementary decision boundaries due to their architectural differences. 
Averaging their predictions effectively reduces prediction variance, a critical advantage in the extreme 5-shot regime.
Let us now focus on how SPARSE\textsubscript{ens} performs with respect to the external competitors. 
The analysis shows that SPARSE\textsubscript{ens} obtains statistically significant improvements over a range of generative models, including SGAN ($p=2.1\times10^{-2},r=1.000$), MatchGAN ($p=4.4\times10^{-2},r=0.636$), CISSL ($p=2.2\times10^{-2},r=0.818$) and EC-GAN ($p=3.5\times10^{-2},r=0.800$). 
The largest performance gap is observed against SGAN, where our method was superior across all 11 datasets. 
The closest competition came from TripleGAN and SEC-GAN; while the differences did not reach statistical significance ($p=1.5\times10^{-1}$ and $p=5.3\times10^{-2}$, respectively), SPARSE\textsubscript{ens} still outperformed them on the majority of datasets with Win-Tie-Loss equal to of 8-0-3 and 9-0-2.
The consistent outperformance of SPARSE\textsubscript{ens} over this diverse set of competitors stems from a crucial methodological distinction. 
All competing methods are purely generative, creating images de novo from a noise vector. 
In contrast, our framework employs image-to-image translation, which modifies existing real, unlabeled images. 
This strategy preserves the authentic and complex anatomical features of the medical data, providing a more robust training signal. 
This fundamental advantage is further amplified when compared to two-player models like SGAN and MatchGAN, which suffer from an internal optimization conflict by tasking a single network with both discrimination and classification. 
Our three-player design avoids this issue. 
Even when compared to advanced three-player models like TripleGAN that also separate these tasks, our method's reliance on translating real images, rather than generating them from noise, appears to be the decisive factor for success in this data-scarce context.

\subsection{Performance in 50-shot}
\label{subsec:50shot}

\begin{table*}[htbp]
    \centering
    \caption{50-Shot statistical comparison of model performance. The upper triangular part shows p-values from statistical comparison (significant values $p < 0.05$ highlighted in bold), while the lower triangular part shows Win-Tie-Loss (W-T-L) statistics. The r-value represents the effect size (Pearson's correlation coefficient) with interpretations: small ($r < 0.3$), medium ($0.3 \leq r < 0.5$), large ($0.5 \leq r < 0.7$), and very large ($r \geq 0.7$). Best performing model is highlighted in bold.}
    \label{tab:50shot_comparison}
    \footnotesize
    \renewcommand{\arraystretch}{1.1}
    \setlength{\tabcolsep}{3.5pt}
    \begin{tabular}{lcccccccc}
    \toprule
    \multirow{2}{*}{Model} & \multicolumn{8}{c}{Statistical Comparison (50 shot)} \\
    \cmidrule(lr){2-9}
    & SPARSE & \textbf{SPARSE\textsubscript{ens}} & SGAN & MatchGAN & CISSL & SEC-GAN & TripleGAN & ECGAN \\
    \midrule
    \multirow{3}{*}{SPARSE} & \multirow{3}{*}{-} & \multirow{3}{*}{\shortstack{1.0e-01\\r=0.636\\(very large)}} & \multirow{3}{*}{\shortstack{\textbf{2.0e-03}\\r=1.000\\(very large)}} & \multirow{3}{*}{\shortstack{\textbf{2.0e-03}\\r=1.000\\(very large)}} & \multirow{3}{*}{\shortstack{\textbf{5.0e-03}\\r=0.818\\(very large)}} & \multirow{3}{*}{\shortstack{\textbf{1.5e-02}\\r=0.636\\(very large)}} & \multirow{3}{*}{\shortstack{3.5e-01\\r=0.455\\(large)}} & \multirow{3}{*}{\shortstack{\textbf{4.0e-03}\\r=0.818\\(very large)}} \\
    \\
    \\
    \multirow{3}{*}{\textbf{SPARSE\textsubscript{ens}}} & \multirow{3}{*}{\shortstack{(9-0-2)}} & \multirow{3}{*}{-} & \multirow{3}{*}{\shortstack{\textbf{2.0e-03}\\r=1.000\\(very large)}} & \multirow{3}{*}{\shortstack{\textbf{2.0e-03}\\r=1.000\\(very large)}} & \multirow{3}{*}{\shortstack{\textbf{2.0e-03}\\r=1.000\\(very large)}} & \multirow{3}{*}{\shortstack{\textbf{2.0e-03}\\r=1.000\\(very large)}} & \multirow{3}{*}{\shortstack{1.0e-01\\r=0.636\\(very large)}} & \multirow{3}{*}{\shortstack{\textbf{2.0e-03}\\r=1.000\\(very large)}} \\
    \\
    \\
    \multirow{3}{*}{SGAN} & \multirow{3}{*}{\shortstack{(0-0-11)}} & \multirow{3}{*}{\shortstack{(0-0-11)}} & \multirow{3}{*}{-} & \multirow{3}{*}{\shortstack{\textbf{8.0e-03}\\r=0.818\\(very large)}} & \multirow{3}{*}{\shortstack{\textbf{2.0e-03}\\r=1.000\\(very large)}} & \multirow{3}{*}{\shortstack{\textbf{2.0e-03}\\r=1.000\\(very large)}} & \multirow{3}{*}{\shortstack{\textbf{2.0e-03}\\r=1.000\\(very large)}} & \multirow{3}{*}{\shortstack{\textbf{1.1e-02}\\r=0.818\\(very large)}} \\
    \\
    \\
    \multirow{3}{*}{MatchGAN} & \multirow{3}{*}{\shortstack{(0-0-11)}} & \multirow{3}{*}{\shortstack{(0-0-11)}} & \multirow{3}{*}{\shortstack{(10-0-1)}} & \multirow{3}{*}{-} & \multirow{3}{*}{\shortstack{9.0e-01\\r=0.091\\(small)}} & \multirow{3}{*}{\shortstack{\textbf{4.0e-03}\\r=0.818\\(very large)}} & \multirow{3}{*}{\shortstack{\textbf{2.0e-03}\\r=1.000\\(very large)}} & \multirow{3}{*}{\shortstack{1.7e-01\\r=0.455\\(large)}} \\
    \\
    \\
    \multirow{3}{*}{CISSL} & \multirow{3}{*}{\shortstack{(1-0-10)}} & \multirow{3}{*}{\shortstack{(0-0-11)}} & \multirow{3}{*}{\shortstack{(11-0-0)}} & \multirow{3}{*}{\shortstack{(6-0-5)}} & \multirow{3}{*}{-} & \multirow{3}{*}{\shortstack{\textbf{1.9e-02}\\r=0.455\\(large)}} & \multirow{3}{*}{\shortstack{\textbf{8.0e-03}\\r=0.636\\(very large)}} & \multirow{3}{*}{\shortstack{5.5e-01\\r=0.091\\(small)}} \\
    \\
    \\
    \multirow{3}{*}{SEC-GAN} & \multirow{3}{*}{\shortstack{(2-0-9)}} & \multirow{3}{*}{\shortstack{(0-0-11)}} & \multirow{3}{*}{\shortstack{(11-0-0)}} & \multirow{3}{*}{\shortstack{(10-0-1)}} & \multirow{3}{*}{\shortstack{(8-0-3)}} & \multirow{3}{*}{-} & \multirow{3}{*}{\shortstack{\textbf{3.1e-02}\\r=0.818\\(very large)}} & \multirow{3}{*}{\shortstack{\textbf{4.0e-03}\\r=0.818\\(very large)}} \\
    \\
    \\
    \multirow{3}{*}{TripleGAN} & \multirow{3}{*}{\shortstack{(3-0-8)}} & \multirow{3}{*}{\shortstack{(2-0-9)}} & \multirow{3}{*}{\shortstack{(11-0-0)}} & \multirow{3}{*}{\shortstack{(11-0-0)}} & \multirow{3}{*}{\shortstack{(9-0-2)}} & \multirow{3}{*}{\shortstack{(10-0-1)}} & \multirow{3}{*}{-} & \multirow{3}{*}{\shortstack{\textbf{2.0e-03}\\r=1.000\\(very large)}} \\
    \\
    \\
    \multirow{3}{*}{ECGAN} & \multirow{3}{*}{\shortstack{(1-0-10)}} & \multirow{3}{*}{\shortstack{(0-0-11)}} & \multirow{3}{*}{\shortstack{(10-0-1)}} & \multirow{3}{*}{\shortstack{(3-0-8)}} & \multirow{3}{*}{\shortstack{(5-0-6)}} & \multirow{3}{*}{\shortstack{(1-0-10)}} & \multirow{3}{*}{\shortstack{(0-0-11)}} & \multirow{3}{*}{-} \\
    \\
    \\
    \bottomrule
    \end{tabular}
\end{table*}

We extend our analysis to the 50-shot setting to examine how model performance scales with increased labeled data availability, while still remaining within the low-labeled data regime.
This configuration provides ten times more labeled samples per class compared to the 5-shot scenario, allowing us to assess the scaling properties of our methods.
Table ~\ref{tab:50shot_comparison} presents the statistical comparison results using the Wilcoxon signed-rank test with Benjamini-Hochberg FDR correction. 
The table follows the same format as the 5-shot analysis, with $p$-values and effect sizes in the upper triangular portion and Win-Tie-Loss statistics in the lower triangular portion. 
The primary finding reveals that SPARSE\textsubscript{ens} maintains its statistical superiority over most competing approaches even as more labeled data becomes available. 
When comparing the two variants of our approach, SPARSE\textsubscript{ens} achieves a Win-Tie-Loss equal to of 9-0-2 against the base SPARSE model. 
However, the $p$-value of $1.0\times10^{-1}$ ($r = 0.636$) indicates no statistical significance: this narrowing gap can be attributed to the changing role of ensemble averaging in different data regimes. 
With only 5 shots per class, individual model predictions are inherently less reliable and more prone to overfitting, making the ensemble approach particularly valuable for reducing prediction variance. 
The discriminator and classifier develop highly complementary decision boundaries due to the scarcity of supervised signals. 
However, with 50 labeled samples per class, the classifier receives sufficient supervision to develop more stable and generalizable representations independently, reducing its reliance on the discriminator's complementary perspective.
Examining SPARSE\textsubscript{ens}'s performance against we note that it achieves statistically significant improvements over SGAN ($p = 2.0\times10^{-3}, r = 1.000$), MatchGAN ($p = 2.0\times10^{-3}, r = 1.000$), CISSL ($p = 2.0\times10^{-3}, r = 1.000$), and EC-GAN ($p = 2.0\times10^{-3}, r = 1.000$), with Win-Tie-Loss equal to 11-0-0 against each. 
These results represent stronger statistical evidence compared to the 5-shot setting, where $p$-values ranged from $2.1\times10^{-2}$ to $4.4\times10^{-2}$. 
Furthermore SPARSE\textsubscript{ens} achieves statistical significance against SEC-GAN ($p = 2.0\times10^{-3}, r = 1.000$, W-T-L: 0-0-11), similarly to what happen in the 5-shot setting. 
When comparing against TripleGAN, we notice that the Win-Tie-Loss ratio is 2-0-9 in favor of our approach, similar to the 5-shot settings, but now the performance differences are not statistically significant at $p=0.1$.
These results suggest that performance differences become more consistent across datasets as labeled data increases.
The patterns observed in this comparison can be attributed to differences in how methods utilize additional supervision. 
Our image-to-image translation approach leverages the increased labeled data to learn more accurate class-conditional transformations. 
With 50 labeled samples per class, the generator receives stronger supervision signals, enabling better preservation of discriminative features during translation. 
Additionally, the quality of pseudo-labels generated in the unsupervised phase improves due to more reliable initial classifiers.

\subsection{Impact of Unsupervised Training Frequency}
\label{subsec:mu}
\begin{figure*}[t]
    \centering
    \includegraphics[width=\textwidth]{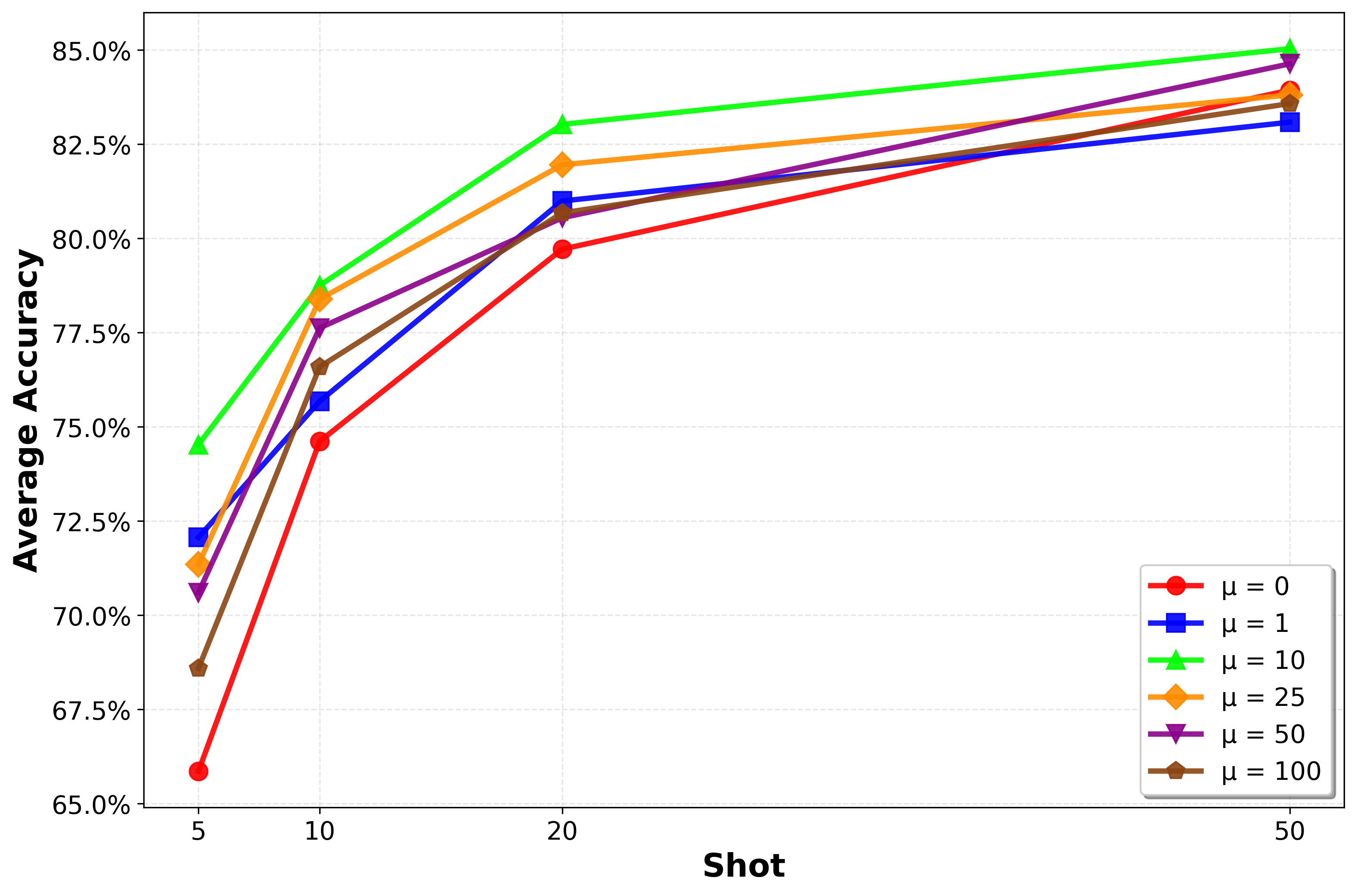}
    \caption{Average classification accuracy as a function of the number of labeled samples per class (shots) for different unsupervised training frequencies ($\mu$). Results are averaged across all eleven MedMNIST datasets and computed on the validation set. The parameter $\mu$ controls how frequently the unsupervised training phase is executed, with $\mu = 0$ representing supervised learning only (no unsupervised phase), $\mu = 1$ indicating unsupervised training at every epoch, and higher values ($\mu \in {10, 25, 50, 100}$) representing unsupervised training every $\mu$ epochs.}
    \label{fig:nu_results}
\end{figure*}
In our approach, the hyperparameter $\mu$ controls how frequently the unsupervised training phase is executed, with the unsupervised phase running every $\mu$ epochs. 
We investigate how varying this parameter affects our model's performance across different few-shot settings, as it represents the balance between supervised and unsupervised learning signals in our semi-supervised framework.
Figure~\ref{fig:nu_results} presents the average accuracy across all datasets as a function of the number of shots per class for different values of $\mu$. These results were obtained using the validation set to avoid any bias that could arise from hyperparameter selection on the test set.
The key finding is that $\mu = 10$ consistently achieves the highest performance across all shot settings. 
In the 5-shot setting, $\mu = 10$ reaches 74.5\% accuracy, representing an 8.5\% point improvement over supervised-only learning ($\mu = 0$ at 66.0\%). 
This performance advantage narrows as labeled data increases, reducing to approximately 2\% points in the 50-shot setting (85.1\% vs 83.1\%).
The figure reveals distinct patterns in how unsupervised learning frequency affects performance. 
Moving from supervised-only learning to any incorporation of unsupervised learning produces improvements, particularly evident in low-shot scenarios. 
Moderate frequencies ($\mu \in \{1, 10, 25\}$) achieve optimal performance, with $\mu = 10$ emerging as optimum. 
Very high frequencies ($\mu \in \{50, 100\}$) lead to performance degradation compared to the optimum, though they still outperform supervised-only learning in data-scarce settings.
These findings validate our design choice of alternating between supervised and unsupervised learning phases, demonstrating that the optimal frequency ($\mu = 10$) remains consistent across different data availability scenarios and that properly scheduled unsupervised learning is particularly crucial in extreme low-data regimes.

\section{Conclusions}
\label{sec:CON}
This paper has introduced a novel GAN-based semi-supervised learning framework specifically designed for extremely low labeled-data regimes. 
Our approach addresses the fundamental challenge of insufficient labeled data through three key innovations: (1) a dynamic training schedule that alternates between supervised and unsupervised phases, (2) an image-to-image translation mechanism that enriches feature representations by learning class-conditional transformations from real unlabeled images, and (3) a confidence-weighted temporal ensemble technique for reliable pseudo-labeling. 
By leveraging these components within a three-player GAN architecture, our method effectively combines the complementary strengths of a generator, discriminator, and dedicated classifier.
The comprehensive empirical evaluation across eleven MedMNIST datasets demonstrates the effectiveness of our approach. In the extreme 5-shot setting, our ensemble configuration 
achieved statistically significant improvements over six state-of-the-art semi-supervised methods, with effect sizes ranging from large to very large. 
The method's superiority stems from its fundamental design choice of performing image-to-image translation rather than generating images from noise, which leverages real medical images as the foundation for learning discriminative features crucial for classification tasks. 
This advantage is further amplified by our temporal ensemble mechanism, which aggregates predictions across training epochs to produce more reliable pseudo-labels in data-scarce scenarios.
Our analysis of the unsupervised training frequency revealed that moderate alternation between supervised and unsupervised phases ($\mu = 10$) balances both learning signals. 
This finding provides practical guidance for deployment, as the optimal frequency remained consistent across different data availability scenarios, eliminating the need for extensive hyperparameter tuning in clinical applications. \\
Despite these promising results, several avenues for future research remain. 
The computational requirements of maintaining multiple networks may pose challenges in resource-constrained clinical settings, motivating the development of more efficient architectures. 
Additionally, extending the framework to incorporate domain-specific medical knowledge and multi-modal imaging data could further enhance its clinical applicability. \\
In conclusion, our GAN-based semi-supervised learning framework represents a significant advancement in addressing the labeled data scarcity challenge in medical imaging. 
By effectively leveraging both labeled and unlabeled data through image translation and ensemble techniques, our method enables robust classification performance even with as few as five labeled samples per class, offering a practical solution for medical imaging applications where annotation costs are prohibitive.

\section*{Acknowledgements}
Guido Manni is a Ph.D. student enrolled in the National Ph.D. in Artificial Intelligence, XXXVII cycle, course on Health and life sciences, organized by Università Campus Bio-Medico di Roma.
This work was partially founded by: i) Piano Nazionale Ripresa e Resilienza (PNRR) - HEAL ITALIA Extended Partnership - SPOKE 2 Cascade Call - "Intelligent Health" with the project BISTOURY - 3D-guided roBotIc Surgery based on advanced navigaTiOn systems and aUgmented viRtual realitY (CUP: J33C22002920006) and ii)  PNRR MUR project PE0000013-FAIR.
Resources are provided by the National Academic Infrastructure for Supercomputing in Sweden (NAISS) and the Swedish National Infrastructure for Computing (SNIC) at Alvis @ C3S.

\bibliographystyle{elsarticle-num}

\bibliography{references}

\clearpage
\FloatBarrier
\appendix

\onecolumn
\section*{Appendix}
\label{app:a}

\begin{center}
Comparison of accuracy scores across different methods on medical image datasets using 5-shot learning. Best performance for each dataset is highlighted in bold.
\label{tab:accuracy_5shot}
\begin{tabular}{lcccccccc}
\toprule
Dataset & SPARSE & SPARSE\textsubscript{ens} & SGAN & MatchGAN & CISSL & SEC-GAN & TripleGAN & ECGAN \\
\midrule
bloodmnist & 0.868 & \textbf{0.887} & 0.173 & 0.302 & 0.463 & 0.813 & 0.862 & 0.241 \\
breastmnist & 0.542 & 0.667 & 0.667 & \textbf{0.792} & 0.615 & 0.719 & 0.677 & 0.667 \\
chestmnist & 0.545 & 0.571 & 0.539 & 0.535 & \textbf{0.610} & 0.533 & 0.510 & 0.532 \\
dermamnist & 0.511 & 0.576 & 0.124 & \textbf{0.626} & 0.566 & 0.535 & 0.563 & 0.626 \\
octmnist & 0.488 & 0.465 & 0.263 & 0.333 & 0.254 & 0.363 & \textbf{0.528} & 0.265 \\
organamnist & 0.780 & \textbf{0.801} & 0.180 & 0.114 & 0.422 & 0.683 & 0.780 & 0.198 \\
organcmnist & 0.754 & \textbf{0.792} & 0.104 & 0.268 & 0.436 & 0.683 & 0.782 & 0.200 \\
organsmnist & 0.567 & \textbf{0.588} & 0.051 & 0.216 & 0.215 & 0.487 & 0.561 & 0.141 \\
pathmnist & 0.751 & \textbf{0.778} & 0.100 & 0.187 & 0.367 & 0.453 & 0.676 & 0.186 \\
pneumoniamnist & 0.796 & 0.806 & 0.585 & 0.710 & 0.785 & \textbf{0.854} & 0.762 & 0.767 \\
tissuemnist & 0.351 & 0.353 & 0.052 & 0.302 & 0.308 & 0.339 & \textbf{0.363} & 0.040 \\
\bottomrule
\end{tabular}
\end{center}

\vspace{2em}

\begin{center}
Comparison of accuracy scores across different methods on medical image datasets using 10-shot learning. Best performance for each dataset is highlighted in bold.
\label{tab:accuracy_10shot}
\begin{tabular}{lcccccccc}
\toprule
Dataset & SPARSE & SPARSE\textsubscript{ens} & SGAN & MatchGAN & CISSL & SEC-GAN & TripleGAN & ECGAN \\
\midrule
bloodmnist & 0.895 & \textbf{0.907} & 0.171 & 0.407 & 0.505 & 0.851 & 0.890 & 0.266 \\
breastmnist & 0.604 & 0.698 & 0.688 & 0.760 & 0.615 & 0.781 & 0.677 & \textbf{0.792} \\
chestmnist & 0.563 & \textbf{0.568} & 0.517 & 0.532 & 0.559 & 0.536 & 0.549 & 0.532 \\
dermamnist & \textbf{0.629} & 0.626 & 0.114 & 0.626 & 0.626 & 0.527 & 0.567 & 0.625 \\
octmnist & 0.490 & 0.515 & 0.246 & 0.240 & 0.316 & 0.481 & \textbf{0.606} & 0.247 \\
organamnist & 0.820 & \textbf{0.834} & 0.194 & 0.108 & 0.393 & 0.795 & 0.797 & 0.103 \\
organcmnist & 0.765 & \textbf{0.808} & 0.148 & 0.744 & 0.405 & 0.676 & 0.793 & 0.167 \\
organsmnist & 0.662 & \textbf{0.694} & 0.050 & 0.495 & 0.230 & 0.521 & 0.605 & 0.188 \\
pathmnist & 0.835 & \textbf{0.873} & 0.099 & 0.372 & 0.332 & 0.657 & 0.792 & 0.050 \\
pneumoniamnist & 0.854 & 0.848 & 0.577 & 0.775 & 0.831 & 0.838 & \textbf{0.881} & 0.802 \\
tissuemnist & 0.417 & \textbf{0.434} & 0.052 & 0.302 & 0.336 & 0.355 & 0.409 & 0.040 \\
\bottomrule
\end{tabular}
\end{center}

\vspace{2em}

\begin{center}
Comparison of accuracy scores across different methods on medical image datasets using 20-shot learning. Best performance for each dataset is highlighted in bold.
\label{tab:accuracy_20shot}
\begin{tabular}{lcccccccc}
\toprule
Dataset & SPARSE & SPARSE\textsubscript{ens} & SGAN & MatchGAN & CISSL & SEC-GAN & TripleGAN & ECGAN \\
\midrule
bloodmnist & 0.925 & \textbf{0.939} & 0.173 & 0.649 & 0.556 & 0.913 & 0.917 & 0.311 \\
breastmnist & 0.698 & \textbf{0.792} & 0.646 & 0.667 & 0.708 & 0.750 & 0.688 & 0.646 \\
chestmnist & 0.563 & 0.574 & 0.525 & 0.562 & 0.574 & 0.565 & \textbf{0.582} & 0.532 \\
dermamnist & 0.652 & \textbf{0.654} & 0.111 & 0.623 & 0.623 & 0.612 & 0.627 & 0.622 \\
octmnist & 0.609 & \textbf{0.667} & 0.248 & 0.334 & 0.292 & 0.463 & 0.642 & 0.259 \\
organamnist & 0.880 & \textbf{0.885} & 0.161 & 0.098 & 0.441 & 0.822 & 0.843 & 0.134 \\
organcmnist & 0.859 & \textbf{0.881} & 0.081 & 0.784 & 0.423 & 0.771 & 0.809 & 0.171 \\
organsmnist & 0.724 & \textbf{0.737} & 0.064 & 0.573 & 0.212 & 0.587 & 0.654 & 0.181 \\
pathmnist & 0.894 & \textbf{0.909} & 0.095 & 0.335 & 0.439 & 0.649 & 0.801 & 0.084 \\
pneumoniamnist & \textbf{0.823} & 0.817 & 0.585 & 0.781 & 0.815 & 0.792 & 0.821 & 0.756 \\
tissuemnist & 0.450 & \textbf{0.472} & 0.052 & 0.302 & 0.326 & 0.440 & 0.470 & 0.075 \\
\bottomrule
\end{tabular}
\end{center}

\vspace{2em}

\begin{center}
Comparison of accuracy scores across different methods on medical image datasets using 50-shot learning. Best performance for each dataset is highlighted in bold.
\label{tab:accuracy_50shot}
\begin{tabular}{lcccccccc}
\toprule
Dataset & SPARSE & SPARSE\textsubscript{ens} & SGAN & MatchGAN & CISSL & SEC-GAN & TripleGAN & ECGAN \\
\midrule
bloodmnist & 0.966 & \textbf{0.968} & 0.172 & 0.819 & 0.612 & 0.936 & 0.956 & 0.807 \\
breastmnist & \textbf{0.844} & 0.813 & 0.646 & 0.708 & 0.698 & 0.781 & 0.792 & 0.750 \\
chestmnist & 0.613 & \textbf{0.616} & 0.522 & 0.532 & 0.604 & 0.568 & 0.577 & 0.532 \\
dermamnist & \textbf{0.708} & 0.701 & 0.133 & 0.625 & 0.619 & 0.617 & 0.660 & 0.624 \\
octmnist & 0.648 & 0.683 & 0.241 & 0.278 & 0.339 & 0.643 & \textbf{0.826} & 0.247 \\
organamnist & 0.928 & \textbf{0.930} & 0.178 & 0.141 & 0.485 & 0.867 & 0.887 & 0.155 \\
organcmnist & 0.884 & \textbf{0.888} & 0.183 & 0.797 & 0.429 & 0.837 & 0.841 & 0.619 \\
organsmnist & 0.746 & \textbf{0.763} & 0.023 & 0.623 & 0.290 & 0.682 & 0.686 & 0.499 \\
pathmnist & 0.906 & \textbf{0.917} & 0.069 & 0.333 & 0.369 & 0.841 & 0.856 & 0.116 \\
pneumoniamnist & 0.813 & \textbf{0.860} & 0.783 & 0.798 & 0.850 & 0.842 & 0.821 & 0.815 \\
tissuemnist & 0.431 & 0.472 & 0.052 & 0.302 & 0.335 & 0.449 & \textbf{0.485} & 0.052 \\
\bottomrule
\end{tabular}
\end{center}

\end{document}